\documentclass[10pt,twocolumn,letterpaper]{article}

\usepackage{cvpr}              

\newcommand{\ytdemtalk}{\href{https://huggingface.co/datasets/anonymous-multimodal-researcher/YT-DemTalk/tree/main}{YT\text{-}DemTalk}}
\usepackage{tikz}
\usetikzlibrary{arrows.meta,positioning,calc,fit,shapes.misc}

\newcommand{\plotW}{0.64\columnwidth} 
\usepackage[font=small,labelfont=bf]{caption}

\usepackage[utf8]{inputenc}
\usepackage{newunicodechar}
\newunicodechar{≈}{\approx}
\usepackage{amsmath,amssymb,mathtools}
\usepackage{algorithm}
\usepackage{algpseudocode}
\algrenewcommand\algorithmicrequire{\textbf{Input:}}
\algrenewcommand\algorithmicensure{\textbf{Output:}}
\usepackage{tabularx,makecell,hhline,booktabs}
\usepackage[table]{xcolor}

\newcolumntype{L}[1]{>{\raggedright\arraybackslash}p{#1}} 
\newcolumntype{C}[1]{>{\centering\arraybackslash}p{#1}}
\newcolumntype{Y}{>{\raggedright\arraybackslash}X}

\usepackage{adjustbox}

\definecolor{cvprblue}{rgb}{0.21,0.49,0.74}
\usepackage[pagebackref,breaklinks,colorlinks,allcolors=cvprblue]{hyperref}

\def\paperID{20036} 
\def\confName{CVPR}
\def\confYear{2026}

\title{Passive Dementia Screening via Facial Temporal Micro-Dynamics Analysis \\of In-the-Wild Talking-Head Video \\}

\author{
Filippo Cenacchi \qquad
Longbing Cao \qquad
Mitchell McEwan \qquad
Deborah Richards\\[0.3em]
School of Computing, Macquarie University, Sydney, Australia\\
{\tt\small filippo.cenacchi@mq.edu.au, longbing.cao@mq.edu.au,}\\
{\tt\small mitchell.mcewan@mq.edu.au, deborah.richards@mq.edu.au}
}

\begin{document}
\maketitle

\begin{abstract}
We target passive dementia screening from short camera-facing talking-head video, developing a facial temporal micro-dynamics analysis for language-free detection of early neuro-cognitive change. This enables unscripted, in-the-wild video analysis at scale to capture natural facial behaviors, transferrable across devices, topics, and cultures without active intervention by clinicians or researchers during recording. Most existing resources prioritize speech or scripted interviews, limiting use outside clinics and coupling predictions to language and transcription. In contrast, we identify and analyze whether temporal facial kinematics, including blink dynamics, small mouth–jaw motions, gaze variability, and subtle head adjustments, are sufficient for dementia screening without speech or text. By stabilizing facial signals, we convert these micro-movements into interpretable facial microdynamic time series, smooth them, and summarize short windows into compact clip-level statistics for screening. Each window is encoded by its activity mix (the relative share of motion across streams), thus the predictor analyzes the distribution of motion across streams rather than its magnitude, making per-channel effects transparent. We also introduce \ytdemtalk{}, a new dataset curated from publicly available, in-the-wild camera-facing videos. It contains 300 clips (150 with self-reported dementia, 150 controls) to test our model and offer a first benchmarking of the corpus. On YT-DemTalk, ablations identify gaze lability and mouth/jaw dynamics as the most informative cues, and light-weighted shallow classifiers could attain a dementia prediction performance of (AUROC) 0.953, 0.961 Average Precision (AP), 0.851 F1-score, and 0.857 accuracy. 
\end{abstract}

\vspace{-4mm}

\section{Introduction}
\label{sec:intro}
Talking-head video has become a reliable substrate for learning fine-grained, in-the-wild audiovisual behavior. CVPR/ICCV work on active-speaker detection and lip reading shows that robust pipelines can extract temporally precise orofacial dynamics under unconstrained conditions \cite{Roth2019AVAACTIVESPEAKER,Chung2017LRS}. In parallel, the community has learned to capture subtle facial micro-motions via dense detection and 3D modeling \cite{Deng2020RetinaFace,Rossler2019FaceForensicspp,Feng2021DECA}. Modern video backbones (Inflated 3D (I3D), R(2+1)D, SlowFast, Video Vision Transformer (ViViT), Video Masked Autoencoders (VideoMAE)) and masked/contrastive pretraining further stabilize temporal representation learning in-the-wild \cite{Carreira2017I3D,Tran2018R2plus1D,Feichtenhofer2019SlowFast,Arnab2021ViViT,Tong2022VideoMAE,He2022MAE}. However, most automated dementia-screening pipelines still anchor on speech/text from constrained interviews (e.g., DementiaBank/Pitt), \cite{dementiabank_talkbank,luz2021adresso,luz2023adressm,yeung2021_alzrt,warnita2018gcnn} which primarily benchmark acoustic–linguistic models under relatively standardized conversational conditions rather than natural, non-clinical settings.  We instead target content-agnostic facial temporal micro-dynamics and blink/eyelid regularity, mouth–jaw motion, gaze dispersion, and head micro-jitter, signals that exist regardless of language or topic. This choice is grounded in neurophysiology: intelligibility and perception rely on slow 1–8\,Hz modulations and on audiovisual coupling between what is heard and what is seen; disruptions in these prosodic-scale dynamics are informative even without lexical content \cite{Giraud2012Cortical,Ding2014Speech,Park2016Lips}. Deployability in-the-wild raises two additional requirements. First, domain transport across channels/devices must work without target labels; shallow, label-free alignment of second-order structure (CORAL/Procrustes) is effective and light-weight \cite{Sun2016DeepCORAL}. Second, well-calibrated probabilities are needed for thresholded screening; temperature scaling remains a strong, simple post-hoc calibrator \cite{Guo2017Calibration}. We handle in-the-wild noise with a face-quality gate that drops unstable frames/windows and summarize the remaining stabilized traces into short-window micro-dynamics scored by a calibrated head. Our formulation focuses on \emph{how} the face moves rather than \emph{what} is said, providing an interpretable and deployable alternative to transcript-based pipelines.

By contrast, \emph{video micro-dynamics} (fine-grained facial/cranial motion and prosodic envelope timing) remain comparatively underexplored despite mounting evidence that temporal facial kinematics and prosodic rhythm jointly encode neurocognitive status. Furthermore, most video-based studies to date have been conducted in clinical or study-managed settings (e.g., scheduled remote conversations), rather than in \emph{in-the-wild} talking-head media comprising publicly available personal vlogs, day-in-the-life recordings, and unstructured interviews, where individuals with or without dementia behave naturally outside experimental contexts, thereby affording greater ecological validity \cite{iconect2024_remoteconv,zhang2025_bhi_face,yeung2021_alzrt,luz2023adressm,park2016_lips,crosse2016_concurrent}.

This gap matters because speech intelligibility and audiovisual perception are governed by \emph{slow modulations} of two coupled signals: (i) the \textbf{acoustic amplitude envelope} of speech and (ii) \textbf{visible articulatory motion} (lips, jaw, eyelids, head). Intelligibility-relevant structure concentrates in the \(\sim\)1–8\,Hz delta–theta band; visual articulators move at similar rates, and their temporal coupling supports robust perception \cite{drullman1994_mods,elliott2009_mtf,giraud2012_cortical,ding2014_continuous,park2016_lips,crosse2016_concurrent}. Disrupting this audio–visual coupling reduces intelligibility even without lexical cues. Consequently, we model the \emph{content-agnostic fluctuation geometry} of facial behavior, the shape and timing of blink recurrence, jaw open–close cycles, and gaze dispersion, rather than the words themselves, reducing sensitivity to language and topic. Complementarily, many physiological and neural signals exhibit \emph{scale-free} structure and long-range temporal dependence, so their variability follows similar patterns across multiple time scales. \emph{Multiscale-entropy}–style measures capture these regularities from short recordings, yielding compact, semantics-free summaries that tend to transfer across cameras, scenes, and recording conditions \cite{he2014_scalefree,costa2002_mse,giraud2012_cortical,ding2014_continuous,elliott2009_mtf}.

Prior human-robot interaction (HRI) studies have already shown the feasibility and usability of \emph{robot-led cognitive assessments} (including Montreal Cognitive Assessment (MoCA)-inspired workflows), suggesting that standardized, unbiased \emph{test administration} and \emph{objective signal capture} are practical in clinics and homes \cite{diNuovo2019_hri_moca,rossi2020_robotics_moca}. Addressing the above gaps and fusing multi-disciplinary approaches, we develop a \emph{content-agnostic}, lightweight, \emph{real-time} detector operating on \emph{facial temporal micro-dynamics} rather than \emph{patient transcripts or demographics}, and we apply this to \emph{humanoid robots} with rich facial expressions (e.g., Pepper/Ameca) as an opportunistic front end for \emph{automated triage and early referral}. This yields an \emph{in-the-wild} pipeline for \emph{passive, real-time} screening from camera-facing video that can be embedded on \emph{humanoids} for proactive diagnosis. To facilitate this unique and reproducible research on passive screening from in-the-wild (naturalistic and variable composition) talking-head video, we curate and release \ytdemtalk{}, a corpus of 300 publicly available interview and monologue recordings. The dataset is balanced: 150 clips are from individuals who explicitly self-report a dementia diagnosis, and 150 clips are from individuals with no such self-report; we use the latter as the control subset. Both groups contain people from a wide range of ages, cultural backgrounds and genders. We provide subject-safe training, validation, and test splits, basic metadata (title, channel identifier, and timestamp) suitable for further open-source testing. Clips with heavy occlusions, severe compression artifacts, or unstable tracking are excluded using a predefined face-quality gate. 

\begin{figure*}[t]
  \centering
  \includegraphics[width=\textwidth]{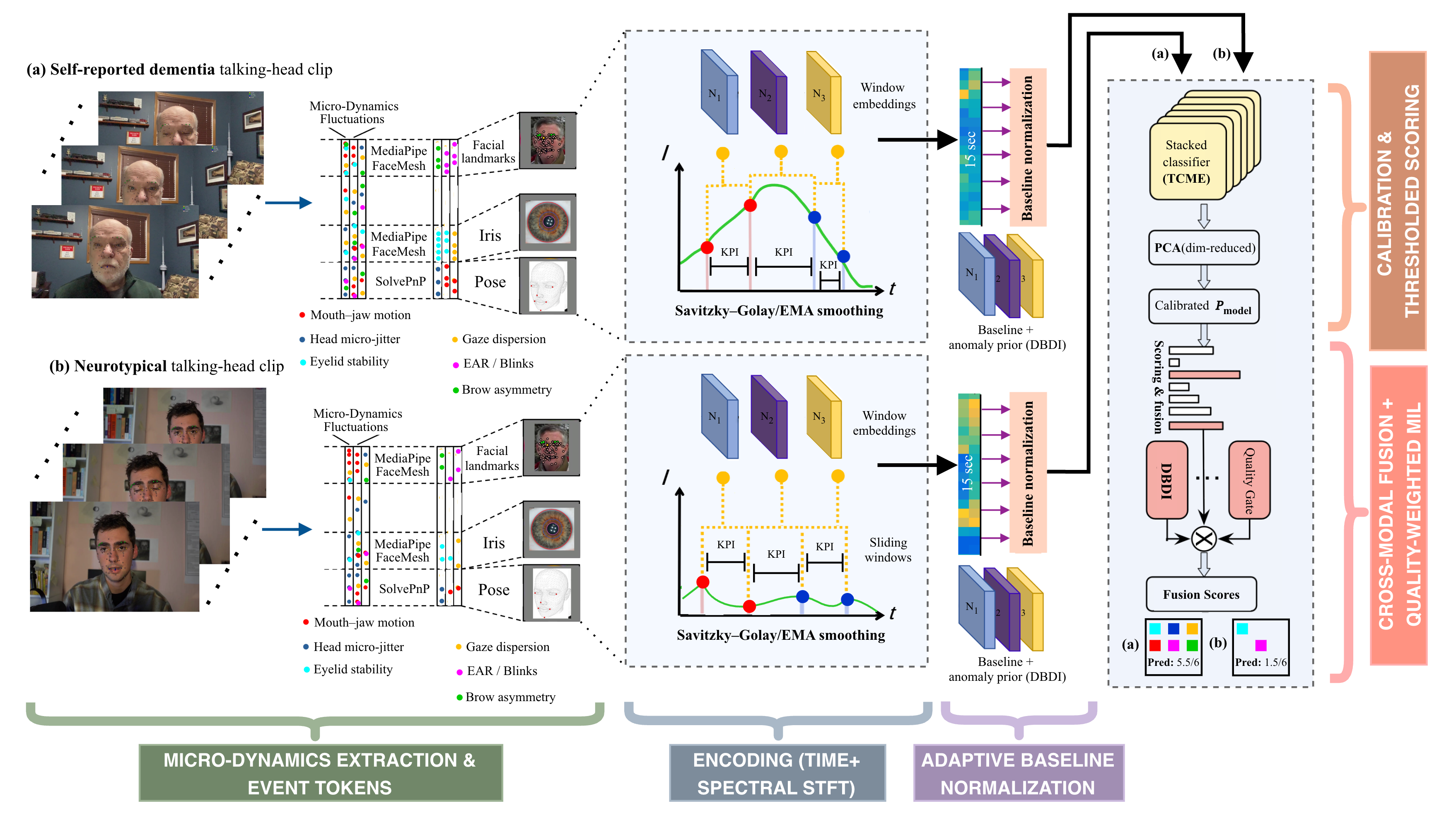}
  \vspace{-2mm}
  \caption{\textbf{Facial micro-dynamics screening pipeline.}
  Frames are stabilized (FaceMesh/Iris with SolvePnP), micro-dynamic key performance indicators (KPIs) are computed and smoothed, 6-s windows (2-s hop) are summarized relative to a 15-s baseline, features are reduced with PCA and scored by a calibrated shallow classifier, and a face-quality gate fuses the scores into one clip-level decision.}
  \label{fig:system32}
  \vspace{-3mm}
\end{figure*}

We study \emph{passive dementia screening from in-the-wild talking-head video}, i.e., single-speaker, self-contained camera-facing clips drawn from public media. Unlike prior work emphasizing curated interviews or speech-only corpora, we target facial micro-dynamics and prosodic envelopes extracted from uncontrolled uploads and evaluate transport across sources  \cite{dementiabank_talkbank,luz2021adresso,iconect2024_remoteconv,zhang2025_bhi_face,yeung2021_alzrt}. Our network (Figure~\ref{fig:system32}; Section~\ref{sec:method}) operates on stabilized facial traces, short-window motion–mix vectors $\mathbf{u}_k\!\in\!\Delta^{5}$ in ILR (Aitchison) space with optional label-free alignment, and a calibrated head that outputs $\hat p(V)$. 
Our work enables scalable, language–free dementia screening from ordinary patients' videos with four contributions:
\begin{itemize}
  \item \emph{Unscripted, content-agnostic facial temporal micro-dynamics.} We operationalize facial micro-dynamics (blink/eyelid regularity, mouth–jaw activity, gaze dispersion, head micro-jitter) for everyday clips, addressing the speech/transcript bias of prior Alzheimer’s disease (AD) resources that rely on controlled audio or automatic speech recognition (ASR) \cite{luz2021adresso}.
  \item \emph{A predictive compositional model in Aitchison geometry.} We treat each window's motion mix \textbf{(6\,s; 2\,s hop)} as a \emph{composition}, map it with an isometric log-ratio (ILR) transform, and calculate distances, covariances, Principal Component Analysis (PCA), and the scoring head directly in the ILR space. We add \emph{reallocation-aware} regularization (Aitchison variance with Dirichlet-style concentration) and \emph{composition-preserving} augmentations (multiplicative noise and stream dropout with renormalization), so the network learns diagnostic \emph{motion reallocation} instead of magnitude shortcuts. 
  \item \emph{Label-free domain alignment in the ILR space.} To transport across channels/devices without target labels, we align source and target ILR statistics using second-order correlation alignment and orthogonal Procrustes, avoiding heavy test-time adaptation while improving robustness in-the-wild \cite{Sun2016DeepCORAL,Wang2021TENT}.
  \item \emph{\textsc{YT-DemTalk} dataset.} We introduce a 300-clip corpus of in-the-wild, camera-facing videos (balanced dementia/control) with subject-safe train/val/test splits to support passive video screening at scale.
\end{itemize}

\section{Related Work}
\label{sec:related}

Large-scale, in-the-wild corpora and tasks around talking heads have driven progress in robust audiovisual modeling. For example, active-speaker detection and diarization on cinematic, unconstrained video (e.g., AVA-ActiveSpeaker) face dynamics with audio cues and stress cross-domain generalization~\cite{Roth2019AVAACTIVESPEAKER,Afouras2018LRS3,Chung2018VoxCeleb2,Chung2017VoxCeleb,Afouras2018LRS2}. Meanwhile, lip-reading and audio-visual speech recognition datasets (LRW/LRS) further emphasize temporal precision of orofacial motion under non-clinical conditions~\cite{Chung2016LRW,Afouras2018LRS2,Afouras2018LRS3}. In parallel, forgery/reenactment benchmarks (FaceForensics++) pressured methods to be sensitive to subtle per-frame micro-dynamics and misalignment~\cite{Rossler2019FaceForensicspp}. By contrast, target a different goal: passive neurocognitive screening where content is incidental and \emph{micro-dynamics} are primary. Consequently, face alignment and 3D morphable modeling have matured rapidly, enabling stable, per-frame geometry for dynamics analysis. Specifically, 3D Dense Face Alignment (3DDFA) solved extreme-pose alignment with a 3D solution~\cite{Zhu2016FaceAlignment3D}, while volumetric CNNs improved single-image 3D reconstruction~\cite{Jackson2017LargePose3D}. Building on this, modern detailed capture (DECA, EMOCA) yields temporally consistent expression parameters usable as micromotor proxies~\cite{Feng2021DECA,Barreto2022EMOCA}. Likewise, real-time reenactment (Face2Face) exposed how tiny expression trajectories drive perceptual identity~\cite{Thies2016Face2Face}. High-quality dense face detectors further stabilized pipelines in-the-wild~\cite{Deng2020RetinaFace}. However, our pipeline leverages these ingredients only as measurement scaffolding; the diagnostic signal is drawn from \emph{content-free} fluctuation geometry. Capturing subtle, multi-scale dynamics from short clips is central to recognition. For temporal modeling, inflated 3D ConvNets (I3D)~\cite{Carreira2017I3D} and factorized 3D convolutions (R(2+1)D)~\cite{Tran2018R2plus1D} showed strong baselines, while SlowFast introduced multi-rate pathways to balance semantics and motion~\cite{Feichtenhofer2019SlowFast}. Lightweight temporal shifting (TSM)~\cite{Lin2019TSM} and transformer-based models (ViViT, VideoMAE)~\cite{Arnab2021ViViT,Tong2022VideoMAE} advanced long-horizon reasoning and pretraining. Forecasting/anticipation tasks further probe whether models internalize causal micro-structure~\cite{Gammulle2019Anticipation,MemFlow2024}. In contrast, our work does not learn high-level actions; instead we quantify \emph{micro-dynamic stability and intermittency} over prosodic time scales and fuse them geometrically. Practically, label scarcity in healthcare motivates self-supervised pretraining. Contrastive and instance-discrimination methods (MoCo)~\cite{He2020MoCo}, vision transformers with masked reconstruction (MAE)~\cite{He2022MAE}, and distillation-based schemes (DINO)~\cite{Caron2021DINO} reduce reliance on dense annotation. For video, masked pretraining transfers well to short clips and preserves motion cues~\cite{Tong2022VideoMAE}. In dementia cohorts, recurring non-verbal motor markers include oculomotor control anomalies (abnormal saccades/fixations/pursuit) \cite{macaskill2016oculomotor}, altered blink/eyelid dynamics \cite{ladas2014blinkmci}, impaired orofacial praxis and jaw–mouth kinematics \cite{cera2013orofacialAD}, changes in facial affect/expressivity (incl. brow movement) \cite{pressman2023facialAD}, and reduced postural stability with increased sway and head micromovement \cite{fernandes2021balanceAD}. For transport, shallow statistical alignment (CORAL/DeepCORAL) ~\cite{Sun2016CORAL} provide effective, label-free matching of second-order structure; adversarial DA complements them when label structure must be preserved~\cite{Tzeng2017ADDA}. For geometry-preserving distribution bridging, optimal-transport variants maintain relational structure~\cite{Thual2022FUGW}. For reliability, deployed screening demands calibrated probabilities. Thus, temperature scaling remains a strong, simple post-hoc calibrator~\cite{Guo2017Calibration}, and detection-oriented objectives like focal loss help under imbalance~\cite{Lin2017Focal}. We combine temperature scaling with Youden-optimal thresholds to keep positive predictive value (PPV) stable across realistic prevalences. CVPR/ICCV have increasingly featured clinically oriented perception systems (e.g., large–small co-modeling for diagnosis~\cite{Chen2025LargeSmall}). Accordingly, our focus is distinct: passive, talking-head \emph{micro-dynamics} as disease-agnostic biomarkers. To our knowledge, no prior work frames passive dementia screening as \emph{cross-modal fluctuation reallocation} measurable from short, in-the-wild clips.


\section{Method}
\label{sec:method}

\vspace{4pt}
\subsection{Preliminary}
\label{subsec:prelim}
We model facial micro-dynamics as a \emph{composition} and learn in composition geometry rather than on raw proportions. From raw frames, we extract dense landmarks and iris centers with MediaPipe FaceMesh/Iris and estimate head pose via SolvePnP to obtain stable, per-frame geometry. Five streams are derived blink/eyelid stability, mouth–jaw motion, gaze dispersion, brow asymmetry, and head micro-jitter, and denoised with Savitzky–Golay and exponential moving averages. Signals are segmented into overlapping windows (6\,s, 2\,s hop) and normalized by a per-video 15\,s baseline to form window embeddings that capture short-horizon behavior. Each window yields a conserved-sum \emph{motion-mix} vector $\mathbf{u}_k\in\Delta^{5}$; we then map to ILR (Aitchison) space, where distances, covariances, and PCA are coherent for compositions. Clip-level features aggregate window means and dispersion in ILR space and are optionally subjected to simple, label-free second-order alignment on the ILR embeddings to improve cross-source transport. A shallow calibrated head maps the reduced features to a reliable posterior $P_{\text{model}}$, and a face-quality gate suppresses low-confidence windows before fusing scores into a clip-level decision. Figure~\ref{fig:system32} illustrates the pipeline on two subjects, one neuro-typical and one with dementia.

\subsection{Rationale}
Passive screening from in-the-wild talking-head video requires signals that (i) are content-agnostic, so language, prompt, and topic do not confound predictions; (ii) remain stable under camera/viewpoint and subject identity; and (iii) are calibratable at inference to support risk thresholds in deployments. Content-dependent baselines (text/semantics, lexical prosody, audio spectra) excel on single streams but entangle diagnosis with “what is said,” microphone conditions, and dataset curation. Purely geometric streams (2D/3D landmarks) can be robust yet lose fine-scale dynamics needed for sensitive triage. Figure~\ref{fig:dashboard} illustrates the resulting system feedback loop. First, a lightweight landmark stabilizer decouples camera shake from facial micromotor activity, yielding per-stream traces that remain interpretable across content. Second, each stream produces a bounded indicator (e.g., blink regularity, eyelid “steadiness,” jaw activity, gaze lability, head micro-movement), which we aggregate into a conserved-sum simplex so no single cue dominates and trade-offs are explicit at thresholding. Third, we prioritize \emph{calibration}: alongside accuracy metrics, we target low Brier score and Expected Calibration Error (ECE) so badge summaries reflect reliable probabilities clinicians, or an automated triage system can act on. This combination, content-agnostic features, conserved compositions, and calibration-first reporting, guides the method described next and underpins our deployment-oriented evaluation.

\vspace{-2pt}

\subsection{Design}
\label{subsec:tasks}

{\setlength{\abovedisplayskip}{4pt}\setlength{\belowdisplayskip}{4pt}\setlength{\abovedisplayshortskip}{3pt}\setlength{\belowdisplayshortskip}{3pt}}

For each video \(V\), we compute per-window micro-dynamics vectors \(\{\mathbf{u}_k\}_{k=1}^{K}\in\Delta^{5}\) (Section~\ref{subsec:prelim}) and summarize them with \(\phi(V)=[\,\bar{\mathbf{u}},\,\mathrm{Disp}(\mathbf{u}_{1{:}K})\,]\in\mathbb{R}^{d}\). A linear head with temperature \(T>0\) yields logit \(\ell=W\phi(V)+b\) and probability \(\hat{p}=\sigma(\ell/T)\). To transport across sources without labels, we estimate an alignment map \(\mathcal{A}\) that matches second-order micro-dynamics structure. Training minimizes a multi-objective: (i) supervised classification on the source domain, (ii) label-free alignment between source/target statistics, and (iii) a composition regularizer that stabilizes the micro-dynamics vectors. 
\vspace{-4pt}
\begin{figure*}[t]
  \centering
  \includegraphics[width=\textwidth]{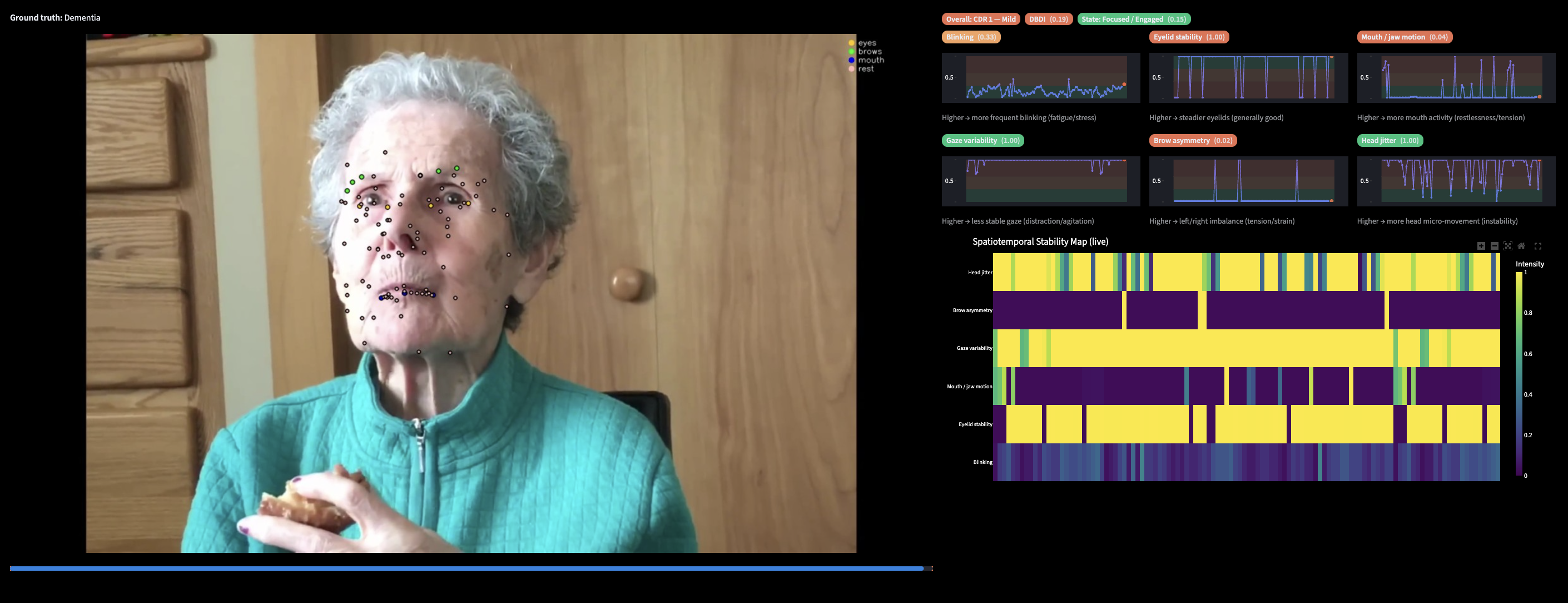}
  \vspace{-6pt}
  \caption{\textbf{Live analysis user interface (UI).} Left: stabilized landmarks over talking-head video. Right: micro-dynamic traces (blink, eyelid stability, mouth/jaw, gaze, head jitter) and a spatiotemporal stability map; top badges summarize calibrated indicators.}
  \label{fig:dashboard}
  \vspace{-6pt}
\end{figure*}

\newcommand{\squeezeeq}{%
  \setlength{\abovedisplayskip}{4pt}%
  \setlength{\belowdisplayskip}{4pt}%
  \setlength{\abovedisplayshortskip}{3pt}%
  \setlength{\belowdisplayshortskip}{3pt}%
}

\squeezeeq
\begin{equation}
\label{eq:loss}
\small
\mathcal{L}
= \frac{1}{|\mathcal{S}|}\!\sum_{(V,y)\in\mathcal{S}}\!
  \mathrm{CE}\!\left(y,\sigma\!\Big(\tfrac{W\phi(V)+b}{T}\Big)\right)
+ \lambda_{\mathrm{aln}}\mathcal{L}_{\mathrm{aln}}
+ \lambda_{\mathrm{reg}}\mathcal{L}_{\mathrm{reg}}.
\end{equation}

This combines supervised classification on compact micro-dynamics vector statistics with label-free source–target alignment and a composition term that stabilizes the simplex representation. For alignment we consider three label-free choices. Let \(\Phi_{\mathcal{S/T}}\) stack \(\phi(\cdot)\) row-wise and \(C(\cdot)\) be an empirical covariance:
\begin{align}
\label{eq:coral}
\small
\mathcal{L}_{\mathrm{coral}}
&= \big\|\,C(\Phi_{\mathcal{S}}) - C(\Phi_{\mathcal{T}})\,\big\|_{F}^{2}
\end{align}
which matches second-order structure and is effective when means are close.
\begin{align}
\label{eq:proc}
\small
\mathcal{L}_{\mathrm{proc}}
&= \min_{R\in\mathbb{R}^{d\times d}}
    \big\|\,\Phi_{\mathcal{S}} - \Phi_{\mathcal{T}} R\,\big\|_{F}^{2}
    \quad \text{s.t.}\; R^{\!\top}R = I
\end{align}
which removes rotation/scale drift via an orthogonal map.
\begin{align}
\label{eq:ot}
\small
\mathcal{L}_{\mathrm{ot}}
&= \min_{\Gamma\in\Pi(\mu_{\mathcal{S}},\mu_{\mathcal{T}})}
     \,\big\langle D,\Gamma\big\rangle
     + \varepsilon\,\mathrm{KL}\!\left(\Gamma \,\middle\|\, \mu_{\mathcal{S}}\mu_{\mathcal{T}}^{\!\top}\right)
\end{align}
which preserves relational geometry when supports only partially overlap using Kullback–Leibler (KL) divergence. To keep the conserved-sum composition and avoid single-cue collapse we add:
  \vspace{-4pt}
\squeezeeq
\begin{equation}
\label{eq:reg}
\small
\mathcal{L}_{\mathrm{reg}}
= \beta\,\mathrm{Var}\!\big[\mathbf{1}^{\!\top}\mathbf{u}_k\big]
+ \gamma \sum_{k}\mathrm{KL}\!\big(\mathbf{u}_k \,\|\, \bar{\mathbf{u}}\big).
\end{equation}

Calibration fits the temperature on validation by negative log-likelihood (NLL):
\squeezeeq
\begin{equation}
\label{eq:nll}
\small
T^{\star} = \arg\min_{T>0}
\sum_{(V,y)\in\mathcal{V}}
\mathrm{CE}\!\left(y,\,\sigma\!\Big(\tfrac{W\phi(V)+b}{T}\Big)\right)
\end{equation}
after which \((W,b,T^{\star})\) are frozen for test-time \(\hat p\).

\begin{algorithm}[t]
\footnotesize
\setlength{\textfloatsep}{6pt}\setlength{\floatsep}{6pt}
\caption{\textbf{Training \& Scoring on Micro-Dynamics} (alignment + calibrated head)}
\label{alg:microdyn}
\begin{algorithmic}[1]
\Require Labeled source $\mathcal{S}$; optional unlabeled target $\mathcal{T}$; weights $\lambda_{\text{aln}},\lambda_{\text{reg}}$; temperature $T$
\Ensure Calibrated scorer $\hat{p}(V)$
\State \textbf{Extract micro-dynamics vectors:} For each $V$, compute $\{\mathbf{u}_k\}$; summarize $\phi(V)=[\,\bar{\mathbf{u}},\mathrm{Disp}(\mathbf{u}_{1:K})\,]$.
\State \textbf{Estimate alignment $\mathcal{A}$ (if $\mathcal{T}$):}
\begin{itemize}
  \item CORAL: minimize $\mathcal{L}_{\text{coral}}$ in Eq.~\eqref{eq:coral};
  \item Procrustes: minimize $\mathcal{L}_{\text{proc}}$ in Eq.~\eqref{eq:proc};
  \item Optimal transport (OT) (entropic/W2): minimize $\mathcal{L}_{\text{ot}}$ in Eq.~\eqref{eq:ot}.
\end{itemize}
If no $\mathcal{T}$, set $\mathcal{A}=\mathbf{I}$.
\State \textbf{Train head:} Minimize Eq.~\eqref{eq:loss} over $(W,b)$ using $(V,y)\!\in\!\mathcal{S}$ with $\phi(V)\!\leftarrow\!\mathcal{A}\phi(V)$.
\State \textbf{Calibrate:} Fit $T^\star$ by NLL (Eq.~\eqref{eq:nll}); freeze $(W,b,T^\star)$.
\State \textbf{Inference:} $\hat{p}=\sigma\!\big((W\,\mathcal{A}\phi(V)+b)/T^\star\big)$; threshold by Youden’s $J$.

\end{algorithmic}
\vspace{-1pt}
\end{algorithm}

\textbf{Micro-dynamics workflow.}
From each input clip we extract facial landmarks, gaze/pose, and (if present) audio traces; we compute per-window micro-dynamics vectors $\{\mathbf{u}_k\}$ (blink, eyelid stability, mouth–jaw motion, gaze variability, head micro-jitter, prosody), summarize each clip with $\phi(V)$, optionally apply label-free alignment, and produce calibrated risk scores. When evaluating across sources without labels, we estimate a label-free alignment
map \( \mathcal{A} \) (CORAL / orthogonal Procrustes / entropic OT) and apply it to \( \phi(V) \) for transport.
A calibrated logistic head then produces the final risk:
\begin{equation}
\hat{p} \;=\; \sigma\!\left(\frac{W\,\mathcal{A}\phi(V)+b}{T}\right),
\end{equation}
where the temperature \(T\) is fit on validation and frozen at test time.
In short, the pipeline performs micro-dynamics extraction, computes optional alignment and adds a calibrated scoring to yield reliable probabilities for screening.

\textbf{Discussion.} The micro-dynamics scoring head optimizes accuracy and reliability jointly: \(\mathcal{L}_{\mathrm{cls}}\) learns a separable decision on compact statistics; \(\mathcal{L}_{\mathrm{aln}}\) grants cross-source transport without labels; and \(T\) ensures operational calibration for thresholded risk. The resulting task formulation is lightweight, semantics-free, and compatible with our live UI (Figure~\ref{fig:dashboard}), contributing to advancements in medical AI.


\section{Experiments and Results}
\label{sec:results}

\subsection{Experimental setup}

Each clip yields \emph{six} micro-dynamics channels: (1) blink/eye aspect ratio (EAR), (2) eyelid stability, (3) mouth–jaw motion, (4) brow asymmetry, (5) gaze dispersion (iris trajectory), and (6) head micro-jitter (pose), computed on sliding windows of 6\,s with a 2\,s hop after a 15\,s per-video baseline. We sweep window/hop, smoothing (Savitzky–Golay vs.\ exponential moving average (EMA)-only), PCA variance targets, and head family. Per-video features (mean and dispersion) are standardized, reduced by PCA (variance target 0.98), and scored by shallow heads (Random Forest (RF), Gradient-Boosted Decision Trees (GBDT), AdaBoost (ADA), Logistic Regression (LR), Support Vector Machine (SVM), Extra Trees (ET); RF is default). For SOTA sequence baselines we additionally train \emph{DL+INCEPTIONTIME} and \emph{DL+TSMIXER} on the same micro-dynamics streams following standard configurations~\cite{fawaz2020inceptiontime,chen2023tsmixer}. Probabilities for all heads are calibrated on validation and frozen for test, following modern large-scale calibration practice in vision~\cite{Minderer2021RevisitingCalibration}. All experiments run on an NVIDIA RTX~4090; full training and evaluation across baselines completes in $\approx$3 hours.

\noindent\textbf{Head (Stats--PCA--RF).}
Here, \emph{Stats} are per-window summary features, \emph{PCA} reduces their dimensionality, and \emph{RF} is a Random Forest trained on the PCA-reduced vectors. We report stratified bootstrap 95\% CIs for AUROC/AP and use non-parametric permutation tests for between-head AUROC differences, which match the ordering in Figs.~\ref{fig:roc_all}--\ref{fig:pr_all}. Because screening acts on thresholds, we calibrate probabilities by temperature scaling on the validation split and keep the temperature fixed at test time~\cite{Minderer2021RevisitingCalibration}. The chosen operating point ($\tau{=}0.636$; Figure~\ref{fig:thr}) lies on a broad Accuracy/F1 plateau and yields recall 0.771, specificity 0.953, accuracy 0.857, F1 0.851, and only two false positives (Figure~\ref{fig:cm_hist}). Single-channel ablations (Figure~\ref{fig:ablate}) show that \emph{gaze} lability and \emph{mouth/jaw} motion contribute most of the signal, with \emph{brow} asymmetry and \emph{head micro-jitter} providing smaller gains; this pattern is stable across the window/hop, smoothing, PCA-target, and head-family sweeps.

\subsection{Results}
\begin{figure}[t]
  \centering
  \includegraphics[width=\plotW]{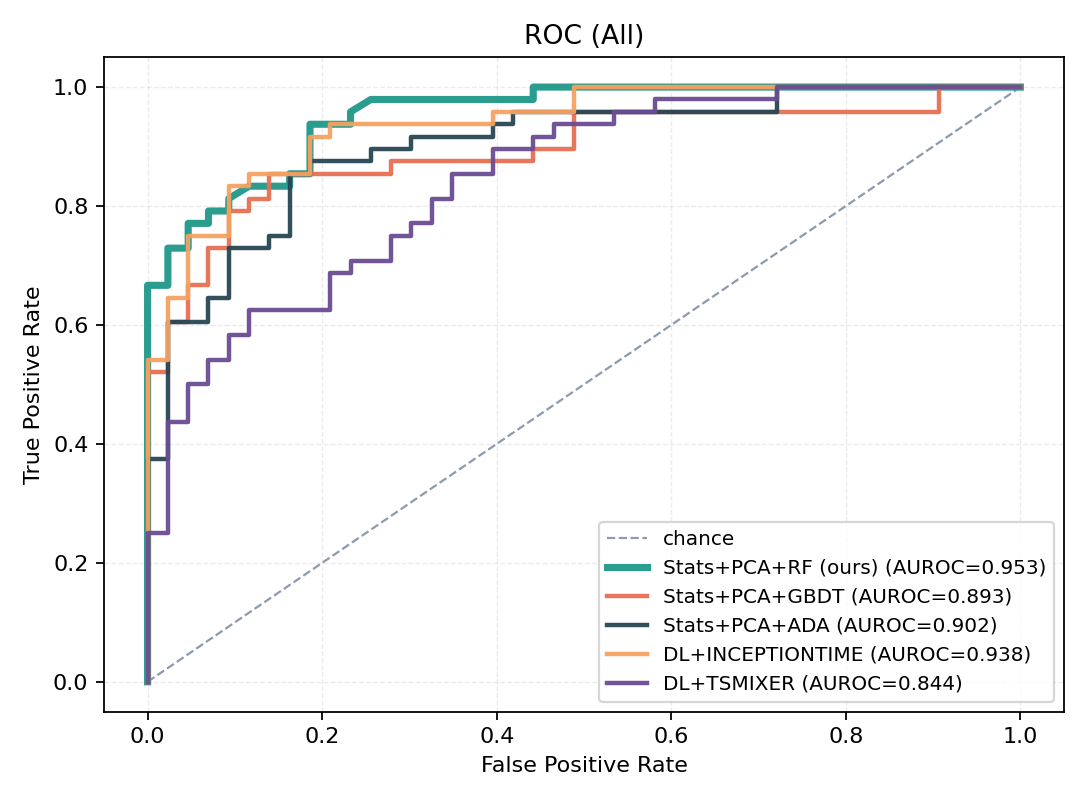}
  \caption{\textbf{ROC (test).} Stats+PCA+RF dominates across most false positive rate (FPR); AUROC=0.953.}
  \label{fig:roc_all}
\end{figure}

\begin{figure}[t]
  \centering
  \includegraphics[width=\plotW]{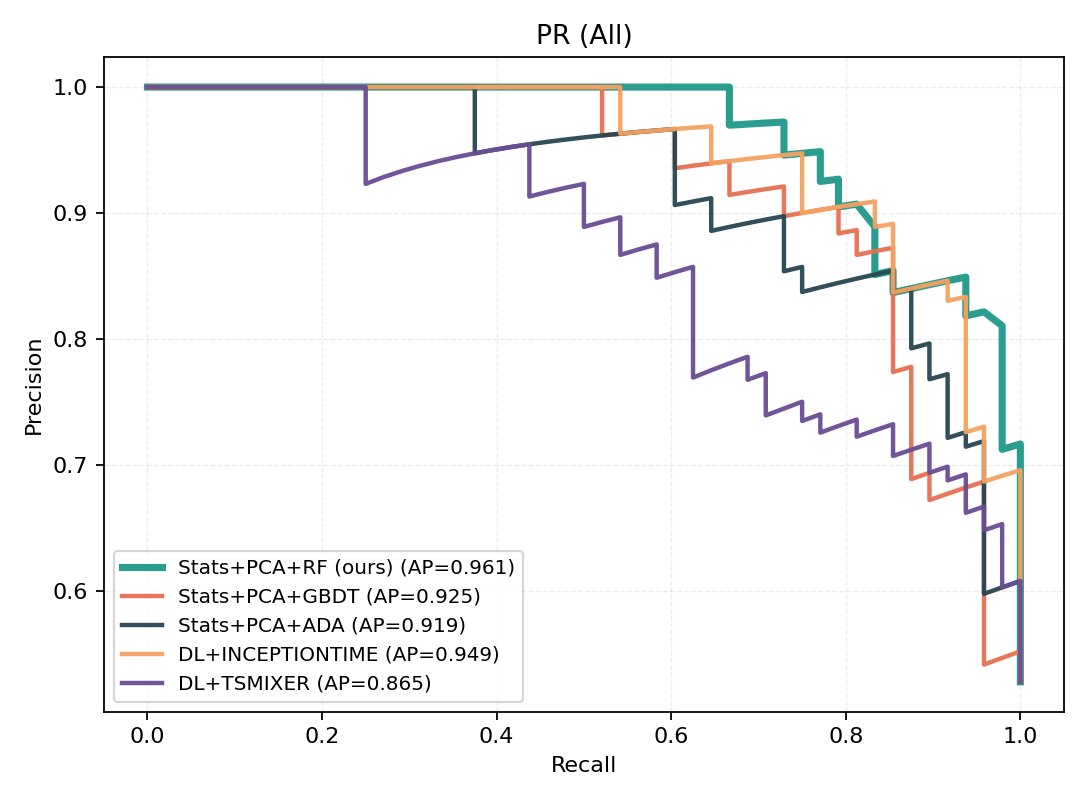}
  \caption{\textbf{Precision–Recall (test).} Near-perfect precision until mid-range recall; AP=0.961.}
  \label{fig:pr_all}
\end{figure}

\begin{figure}[t]
  \centering
  \begin{minipage}{0.49\linewidth}
    \centering
    \includegraphics[width=\linewidth]{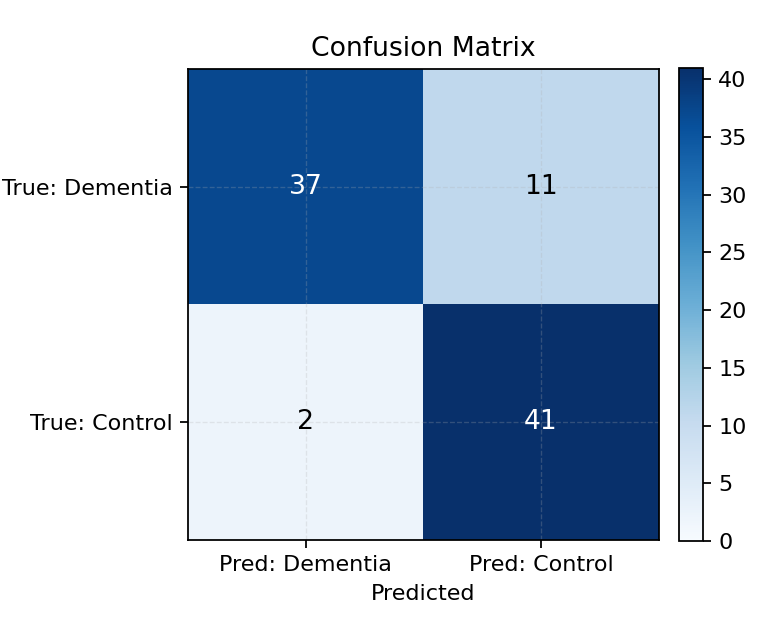}
  \end{minipage}\hfill
  \begin{minipage}{0.49\linewidth}
    \centering
    \includegraphics[width=\linewidth]{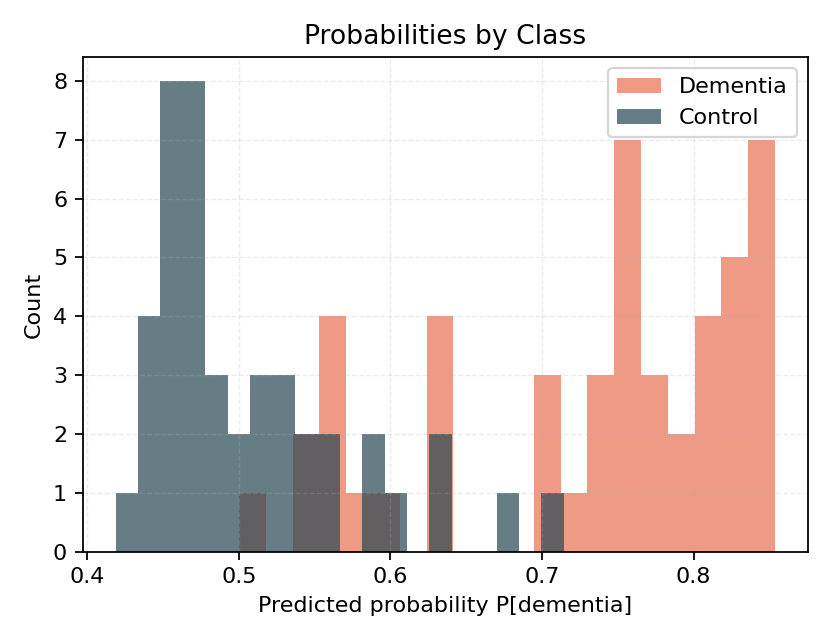}
  \end{minipage}
  \vspace{-6pt}
  \caption{\textbf{Left:} Confusion matrix at the validation–tuned threshold ($\tau{=}0.636$). \textbf{Right:} Class-conditioned probability histograms show good separation with a narrow overlap band.}
  \label{fig:cm_hist}
  \vspace{-2pt}
\end{figure}

\begin{figure}[t]
  \centering
  \includegraphics[width=\plotW]{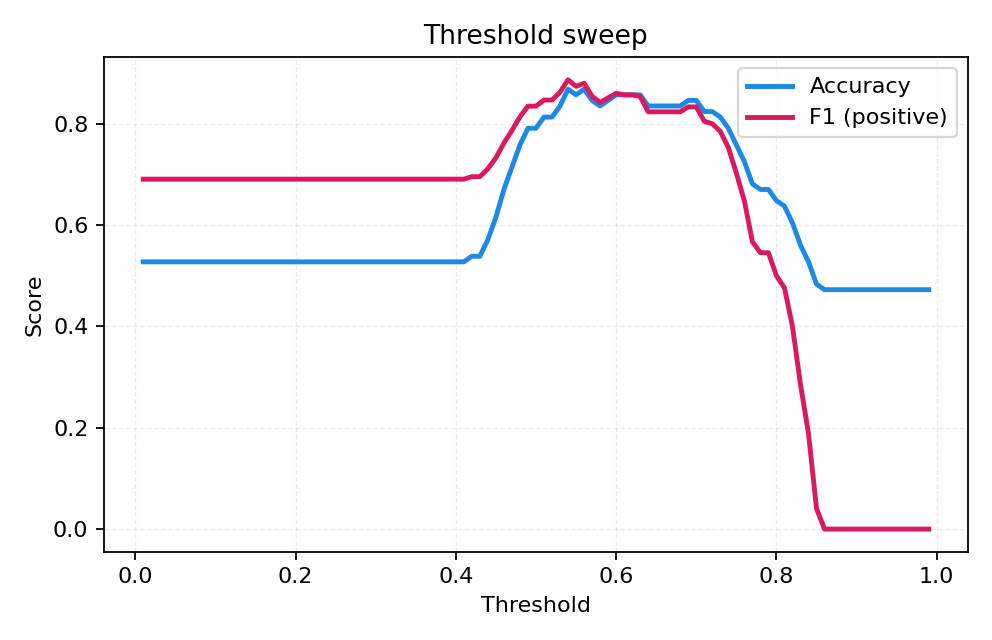}
  \caption{\textbf{Threshold sweep.} A broad optimum exists; $\tau{=}0.636$ balances precision and recall with slight bias toward higher PPV.}  \label{fig:thr}
  \vspace{-12pt}
\end{figure}

\begin{figure}[t]
  \centering
  \includegraphics[width=\plotW]{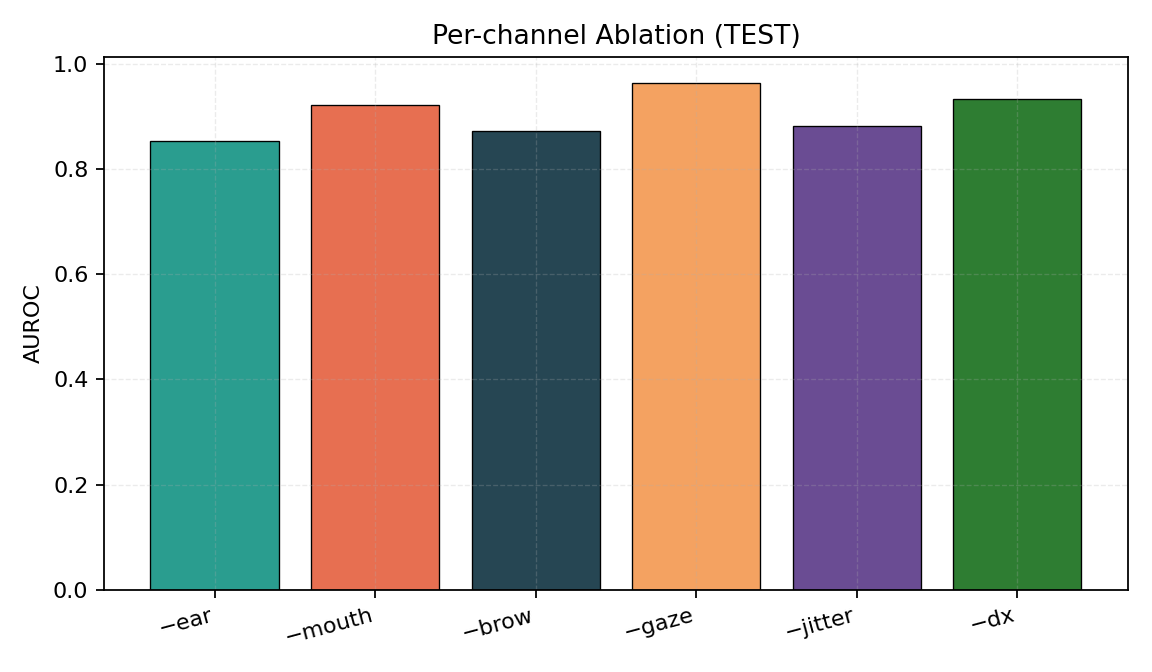}
  \vspace{-6pt}
  \caption{\textbf{Per-channel ablation (test AUROC).} Removing \emph{gaze} or \emph{mouth/jaw} hurts most; other channels contribute but are less critical by themselves.}
  \label{fig:ablate}
  \vspace{-8pt}
\end{figure}

\begin{figure*}[t]
  \centering
  \includegraphics[width=.96\textwidth]{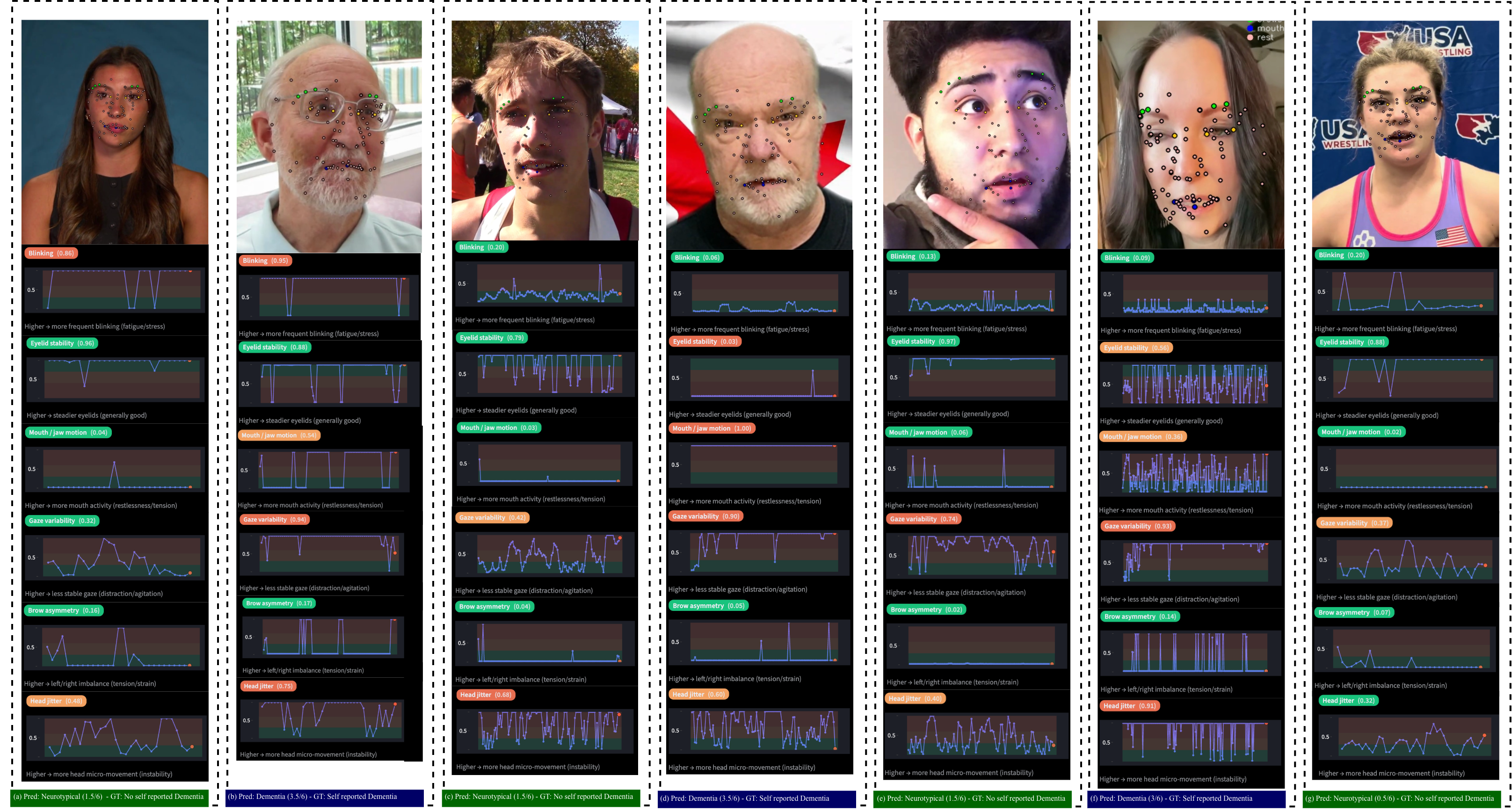}
  \vspace{-6pt}
  \caption{\textbf{Qualitative examples.} Stabilized face mesh (top) with synchronized micro-dynamics traces (bottom). Badges summarize calibrated indicators.}
  \label{fig:qual}
  \vspace{-6pt}
\end{figure*}

Table~\ref{tab:perf_new} summarizes the headline discrimination numbers for five heads, contrasting our Stats+PCA+RF default against a convolutional SOTA baseline (DL+INCEPTIONTIME), an MLP mixer (DL+TSMIXER), and two stronger tree ensembles (Stats+PCA+ADA/GBDT). The Stats+PCA+RF head attains the best AUROC, AP, F1 and Acc, so we keep it as our primary screening model for all subsequent analyses. The next figures unpack these results in a consistent order. First, Figures~\ref{fig:roc_all} and~\ref{fig:pr_all} focus on ranking quality under different prevalence and threshold conditions. Second, Figure~\ref{fig:cm_hist} connects the selected operating point to concrete error types and the distribution of posteriors. Third, Figure~\ref{fig:thr} shows how metrics change as the decision threshold moves, which clarifies the trade-off between precision and recall. Fourth, Figure~\ref{fig:ablate} examines the marginal contribution of each microdynamic channel. Finally, Figure~\ref{fig:qual} provides qualitative examples that make quantitative patterns visible and interpretable.

\begin{table}[H]
\centering
\footnotesize
\setlength{\tabcolsep}{6pt}
\begin{tabular}{lccccc}
\toprule
Method & AUROC & AP & F1 & Acc & Thr \\
\midrule
\textbf{Stats+PCA+RF (ours)} & \textbf{0.953} & \textbf{0.961} & \textbf{0.851} & \textbf{0.857} & 0.636 \\
DL+INCEPTIONTIME            & 0.938 & 0.949 & 0.824 & 0.835 & \textbf{0.794} \\
Stats+PCA+ADA               & 0.902 & 0.919 & 0.741 & 0.769 & 0.690 \\
Stats+PCA+GBDT              & 0.893 & 0.925 & 0.844 & 0.846 & 0.522 \\
DL+TSMIXER                  & 0.844 & 0.865 & 0.772 & 0.747 & 0.662 \\
\bottomrule
\end{tabular}
\caption{Test performance with validation-tuned thresholds.}
\label{tab:perf_new}
\end{table}
\vspace{-3pt}

We benchmark our Stats+PCA+RF head against InceptionTime as a strong convolutional TSC reference~\cite{fawaz2020inceptiontime}, PatchTST and iTransformer as channel-aware Transformer forecasters~\cite{nie2023patchtst,itransformer2024}, TimesNet as a 2D-variation temporal model~\cite{wu2023timesnet}, and TSMixer as an efficient all-MLP mixer~\cite{chen2023tsmixer}, so that our calibrated tree ensemble is tested against diverse, state-of-the-art (SOTA) inductive biases. Figure~\ref{fig:roc_all} summarizes ranking performance across all possible thresholds. The curve for the Stats, PCA, then Random Forest head dominates the alternatives over a wide false positive range, which supports the aggregate AUROC reported in Table~\ref{tab:perf_new}. This indicates that the chosen representation and head separate positive and negative clips consistently, even when the operating point is shifted. The shape near the low false positive region is especially relevant for screening, where the cost of false positives is high and precision must remain stable.

\textbf{Interpreting Precision–Recall.}
Figure~\ref{fig:pr_all} complements the ROC view by stressing behavior when positive cases are relatively rare. The area under the precision–recall curve remains high, and precision stays near perfect until mid-range recall. This is aligned with the screening goal stated in the setup, which favors a conservative operating point that keeps positive predictive value high while still recalling a substantial fraction of true positives.

\textbf{Linking the operating point to errors and posteriors.}
Figure~\ref{fig:cm_hist} ties the threshold $\tau{=}0.636$ to concrete outcomes. The left panel shows that false positives are few, which is consistent with the conservative choice of threshold. The right panel shows class-conditioned probability histograms with a visible separation band and a relatively narrow overlap region. Together, these two views explain why the selected threshold yields high precision while keeping recall at a workable level for screening.

\textbf{Operating-point selection.}
Figure~\ref{fig:thr} shows that accuracy and F1 have a broad plateau around the chosen threshold. This plateau suggests that small deviations during deployment, for example due to mild distribution shift, will not degrade performance sharply. The selected operating point therefore reflects a practical balance for real-time use, where the system must remain reliable without per-site retuning. Figure~\ref{fig:ablate} quantifies the marginal impact of each stream by ablating them one at a time. The largest drops occur when gaze variability or mouth and jaw motion are removed, while brow asymmetry and head micro-jitter have smaller, but still positive, effects. This pattern matches the intended design of the micro-dynamics representation: orofacial and oculomotor dynamics carry stable, content-agnostic cues in ordinary talking-head video, and they anchor the performance of the calibrated head.

\textbf{Results}
We report AUROC and AP with stratified bootstrap 95\% CIs, and calibration with Expected Calibration Error (ECE; 10 bins) and Brier score. Between-method AUROC differences are assessed via a non-parametric, label-preserving permutation test, following recent best practice for calibration/uncertainty evaluation and hypothesis testing in vision \cite{cheng2022pairwise,zhang2023pitfall,mandel2024permutation}. Our chosen head, \emph{Stats+PCA+RF (ours)}, attains
\textbf{AUROC}~0.953\,\([0.912,\,0.984]\) and
\textbf{AP}~0.961\,\([0.928,\,0.986]\),
with \textbf{F1}~0.851 and \textbf{ACC}~0.857 at the tuned threshold~0.636.
Calibration is reasonable: \textbf{ECE}~0.268 and \textbf{Brier}~0.168.
Among the new baselines, the strongest competitor is the convolutional \emph{DL+INCEPTIONTIME} head, which reaches AUROC~0.938 and AP~0.949 (F1~0.824, ACC~0.835), while the best alternative tree model, \emph{Stats+PCA+GBDT}, attains AUROC~0.893, AP~0.925, F1~0.844 and ACC~0.846, and \emph{DL+TSMIXER} lags further behind (AUROC~0.844, AP~0.865). Permutation tests on AUROC differences find no significant gap between \emph{Stats+PCA+RF} and the next-best heads, but all alternatives show higher calibration error, so \emph{Stats+PCA+RF} remains the best compromise between discrimination and reliability under subject-safe splitting.

\section{Analysis}
\label{sec:analysis}
\subsection{Qualitative Results}
\label{sec:qualitative}
We examine how the model behaves beyond headline metrics. We first examine qualitative evidence on held-out clips, then connect the selected operating point to calibration, and finally analyze which micro-dynamics channels matter most and where errors arise. Figure~\ref{fig:qual} shows stabilized landmarks overlaid on talking-head frames with synchronized traces for blink/eyelid, mouth–jaw, gaze, brow asymmetry, and head micro-jitter. Clips with sustained \emph{gaze lability} and elevated \emph{mouth–jaw} activity receive higher posteriors, whereas clips with steady gaze and moderate articulatory motion are scored as low risk. The visible cues align with the ranking behavior seen in ROC/PR views (Figs.~\ref{fig:roc_all}–\ref{fig:pr_all}) and with the separation of class-conditioned posteriors in Figure~\ref{fig:cm_hist}. The UI exposes per-stream indicators and a calibrated score badge. Clinicians can verify that risk increases when gaze wanders during speech or when jaw cycles become irregular, and decreases when the face is well-framed with stable ocular behavior.

\subsection{Operating Point and Calibration}
\label{sec:calibration}
The validation-tuned threshold $\tau{=}0.636$ lies on a broad plateau for Accuracy/F1 (Figure~\ref{fig:thr}), indicating tolerance to mild distribution shift. At this point, the test confusion matrix shows few false positives and a clear separation band in the posterior histograms (Figure~\ref{fig:cm_hist}). Calibration is reasonable (ECE 0.268; Brier 0.168), so reported probabilities are interpretable for screening rather than only for ranking. Single-channel removals (Figure~\ref{fig:ablate}) identify \emph{gaze} variability and \emph{mouth–jaw} motion as the dominant contributors; ablating either causes the largest AUROC drops. \emph{Brow} asymmetry and \emph{head micro-jitter} provide secondary gains, suggesting that orofacial and oculomotor streams anchor performance in uncontrolled video. Common failures mirror these findings. False negatives concentrate in clips with (i) mouth occlusion or beard-induced tracking gaps, (ii) extreme head pose or framing that suppresses iris tracking, and (iii) compression artifacts that destabilize high-frequency micromovements. Simple capture guidance camera-facing framing, moderate lighting, and minimal mouth occlusion reduces these errors and preserves the same qualitative cues surfaced in the UI.

\section{Conclusion}
We present a content-agnostic, micro-dynamics approach to passive dementia screening from ordinary talking-head videos. The proposed micro-dynamics representation, built from stabilized facial traces, short-window statistics, and a conserved composition, pairs naturally with label-free alignment (CORAL/Procrustes) and post-hoc calibration. On our curated data \ytdemtalk{}, the compact PCA\,then\,RF head achieves strong discrimination (AUROC~0.953, AP~0.961) and high precision at a conservative operating point, while ablations show that gaze lability and mouth/jaw motion carry the largest marginal signal. The live analysis UI exposes per-stream indicators and calibrated scores, yielding traceable, per-stream evidence. Our study is limited to a single speaker, short clips without assessing longitudinal change, clinic-grade labels, or fairness across demographic strata. Future work will: (i) couple the micro-dynamics representation with self-supervised video pretraining for stronger cross-domain transport, (ii) extend to multi-view and longer-horizon stability measures, and (iii) embed the larger data with deeper models on resource-constrained \emph{humanoid robots} for on-device, in-the-wild triage with human-in-the-loop oversight, evaluating usability and calibration drift under real deployment.

{
    \small
    \bibliographystyle{ieeenat_fullname}
    \bibliography{main}
}


\end{document}